\documentclass[letterpaper, 10 pt, conference]{ieeeconf}  

\IEEEoverridecommandlockouts                              

\usepackage{graphicx, tipa}
\usepackage{amsmath,amssymb,amsfonts}
\usepackage[caption=false]{subfig}
\usepackage[ruled,linesnumbered, noend]{algorithm2e}
\usepackage{dirtytalk}
\usepackage{hyperref}
\usepackage{gensymb}
\usepackage{tabularx}
\usepackage{multirow}
\usepackage[short]{optidef}
\usepackage{booktabs}
\usepackage{siunitx}
\usepackage{titlesec}
\usepackage[dvipsnames]{xcolor}
\usepackage{colortbl}   

\newcounter{cases}
\newcounter{subcases}[cases]

\makeatletter
\newcommand{\removelatexerror}{\let\@latex@error\@gobble}
\makeatother
\SetKwComment{Comment}{$\triangleright$\ }{}

\newcommand\Tstrut{\rule{0pt}{2.0ex}}         

\title{\huge 
VL-DPO: Vision-Language-Guided Finetuning for Preference-Aligned Autonomous Driving
}

\author{Zhefan Xu$^{1}$, Ghassen Jerfel$^{2}$,  Marina Haliem$^{2}$, Qi Zhao$^{2}$, Jeonhyung Kang$^{2}$, and Khaled S. Refaat$^{2}$ 
\thanks{$^{1}$ {\small zhefanx@andrew.cmu.edu}}%
\thanks{$^{2}$ {\small \{ghassen,mhaliem,zhaqi,jeonhyung,krefaat\}@waymo.com}}%
}

\begin{document}

\maketitle
\thispagestyle{empty}
\pagestyle{empty}

\begin{abstract}
The rapid growth of autonomous driving datasets has enabled the scaling of powerful motion forecasting models. While large-scale pretraining provides strong performance, the standard imitation objective may not fully capture the complex nuances of human driving preferences. Meanwhile, recent advances in vision-language models (VLMs) have demonstrated impressive reasoning and commonsense understanding.
Building on these capabilities, this paper presents VL-DPO, a vision-language-guided framework that aligns ego-vehicle motion forecasting models with human preferences. Our approach leverages a VLM as a zero-shot reasoner to automatically generate preference pairs from a pretrained model's rollouts, which are then used to finetune the model via Direct Preference Optimization (DPO). 
We finetune our models on the Waymo Open End-to-End Driving Dataset (WOD-E2E) and evaluate performance against held-out human preference annotations using rater feedback score (RFS) and average displacement error (ADE). Our experiments confirm that the VLM's trajectory selection is a high-quality proxy for human preference. Our final model, VL-DPO, yields an 11.94\% increase in RFS and a 10.01\% reduction in ADE over the pretrained model. 
\end{abstract}

\section{Introduction}
Deep neural networks, empowered by the proliferation of large-scale datasets, have demonstrated strong performance in motion forecasting for autonomous driving \cite{huang2024drivegpt}\cite{motionlm}\cite{ wayformer}\cite{MultiPath++}\cite{baniodeh2025scaling}. Despite their effectiveness, the standard imitation learning objectives prioritize local geometric accuracy, often via next-token prediction accuracy. This fails to capture the holistic preferences that characterize human driving behavior, creating an alignment gap.

The recent emergence of Vision-Language Models (VLMs), with their strong commonsense reasoning and contextual understanding, offers a promising path to bridge this gap. We hypothesize that a VLM's vast world knowledge can serve as a high-quality proxy for nuanced human preferences.

However, the prevailing trend for leveraging this capability has been to adopt VLMs as monolithic end-to-end backbones for ego-vehicle motion prediction \cite{DriveLM}\cite{DriveVLM}\cite{WiseAD}\cite{EMMA}\cite{OpenDriveVLA}. While these approaches benefit from the large-scale pretrained knowledge embedded in VLMs and have shown promising results, they still face several key challenges. First, these models often require large-scale, high-quality, and diverse datasets with aligned language and action (as exemplified by the data curation process in \cite{RT2}), which can be expensive and time-consuming to collect and annotate. Second, without careful data curation, finetuning large pretrained models on domain-specific driving datasets can lead to catastrophic forgetting \cite{InstructVLA}, eroding the general world knowledge that initially underpins their effectiveness \cite{ChatVLA}. Third, the black-box nature of these end-to-end models makes it difficult to interpret individual decisions, posing a critical barrier for safety-sensitive applications such as autonomous driving. Fourth, the computational demands of large VLMs often lead to high inference latency, making them impractical for real-time onboard deployment in autonomous vehicles.

\begin{figure}[t] 
    \centering
    \includegraphics[scale=0.26]{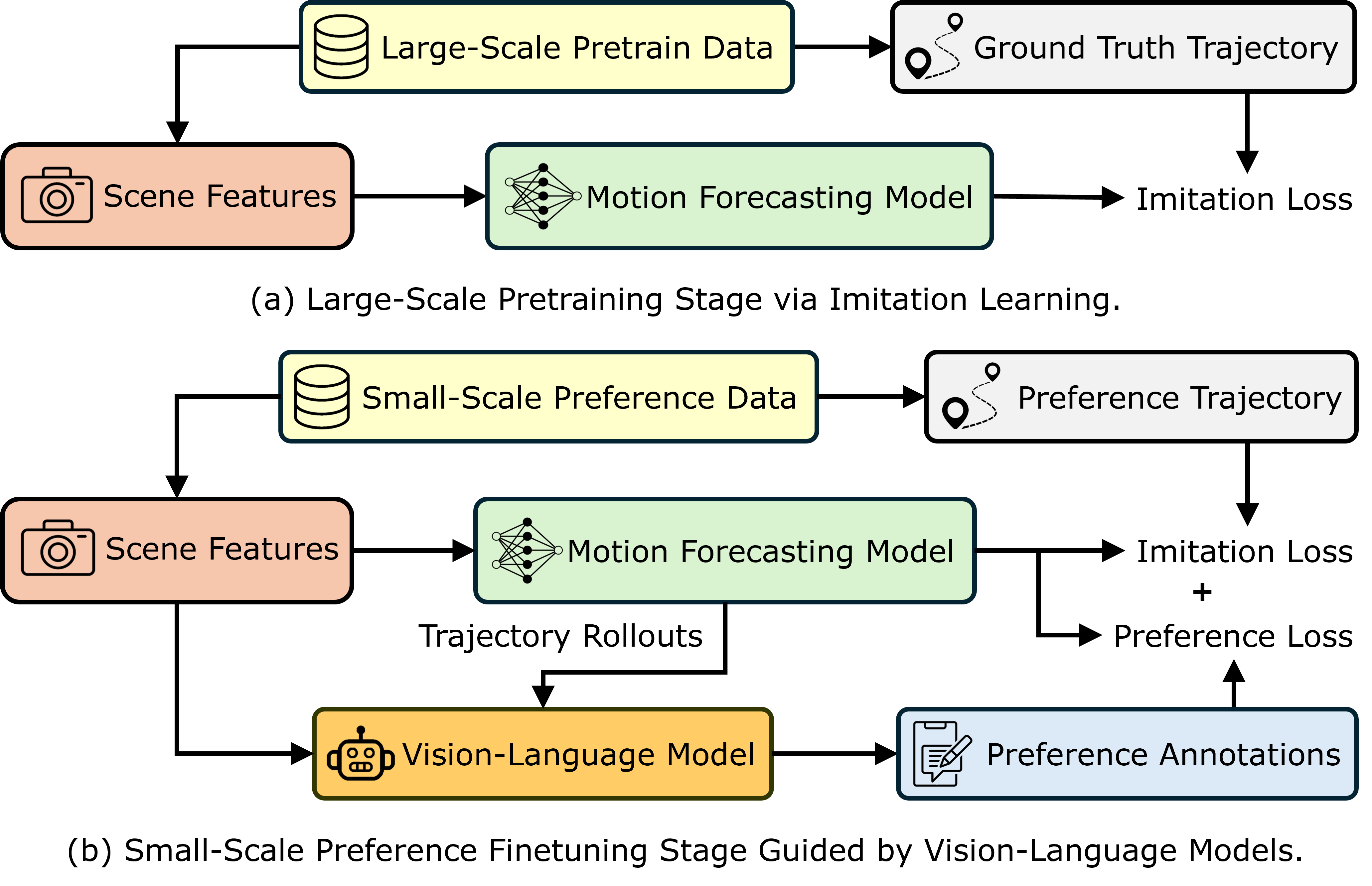}
    \caption{Illustration of the proposed training stages of the motion forecasting model. (a) Multi-agent pretraining via imitation learning on large-scale data. (b) Single-agent (ego-vehicle) preference finetuning using VLM-generated pairs, which replace human annotation.}
    \label{fig:intro_figure}
\end{figure}

To address the limitations of finetuning VLMs as Vision-Language-Action models, we propose a more modular paradigm that decouples high-level reasoning from low-level motion forecasting by leveraging the VLM not as an integrated component, but as a frozen zero-shot reasoner and critic. This design preserves the VLM's world knowledge and avoids catastrophic forgetting while enabling a more interpretable and efficient finetuning process. In this role, the VLM generates supervisory signals, including fine-grained trajectory preference pairs—our main contribution—and coarser discrete High-Level Actions (HLAs), which we also investigate as a simpler supervision method.


This culminates in VL-DPO: a vision-language-guided framework that aligns a pretrained motion forecasting model with human preferences, as illustrated in Fig. \ref{fig:intro_figure}. The framework builds on Direct Preference Optimization (DPO) \cite{DPO}, where the VLM evaluates candidate trajectories from model rollouts to automatically generate preference pairs for finetuning. Our empirical study on the Waymo Open End-to-End Driving Dataset (WOD-E2E) shows that VL-DPO consistently outperforms the coarser HLA-based supervision and achieves state-of-the-art alignment with human preferences. The main contributions of this work are:

\begin{itemize}
    \item \textbf{VLM as a Zero-Shot Preference Annotator:} We propose the novel use of a VLM as a zero-shot reasoner to generate preference pairs for supervising a separate motion forecasting model. We validate this approach by demonstrating that the VLM's trajectory selections are more human-aligned than the model's own most-likely predictions, establishing it as an effective and scalable proxy for human feedback.
    \item \textbf{VL-DPO for Motion Forecasting:} We present the first framework that leverages vision-language-guided preferences within direct preference optimization to significantly improve the alignment of ego-vehicle motion forecasting models with human driving preferences.
    \item \textbf{Comprehensive Empirical Study:} 
    On WOD-E2E, our proposed VL-DPO framework demonstrates significant improvements over both a pretrained model and a strong imitation learning baseline (finetuned on the single highest-rated human preference trajectory), while consistently outperforming alternative VLM-guidance strategies like HLA-based supervision.
\end{itemize}

The rest of this paper is structured as follows: we review related work in Section~\ref{sec:related_works}, detail our methodology in Section~\ref{sec:methodology}, present our experimental results in Section~\ref{sec:results}, and finally conclude with a discussion of our findings.
\section{Related Work}
\label{sec:related_works}
Motion forecasting, crucial for autonomous driving, has advanced from CNNs on rasterized scenes \cite{mtp}\cite{rules_of_road}\cite{multipath}, and GNNs on scene graphs \cite{spagnn}\cite{lanegcn}, to powerful Transformer-based architectures \cite{wayformer}\cite{MTR}\cite{hdgt}. 
Paradigms like MotionLM \cite{motionlm}, reformulate it as a language modeling for motion tokens, effectively capturing complex interactions and multimodal motion distributions. While achieving high accuracy, its standard next-token imitation training objective is primarily optimized for replicating expert-demonstrated trajectories. This focus on local fidelity means it remains misaligned with the long-term consistency and holistic nature of human driving behavior, as it does not explicitly account for nuanced preferences or explore alternative, equally valid or even preferred paths that may deviate from the exact expert trace. This motivates our work on human preference alignment in motion forecasting.

\textbf{VLM-based Approaches:} The dominant approach in leveraging VLMs for autonomous driving involves finetuning them as monolithic end-to-end models \cite{e2e_multi-modal}\cite{VAD}\cite{UniAD}; for instance, DriveLM \cite{DriveLM} trains a VLM for trajectory prediction via a VQA framework, WiseAD \cite{WiseAD} enriches this with explicit driving knowledge, and EMMA \cite{EMMA} finetunes a VLM to output both reasoning traces and trajectory predictions. Others, like VDT-Auto \cite{VDT-Auto}, use a VLM to output embeddings for a policy module while DiMa \cite{DiMa} relies on the joint-training of a VLM and a vision-based policy model. All of these approaches share the common trait of directly modifying the VLM's weights for the driving task which can erode its world knowledge and reasoning capabilities.

In contrast, our work adopts the less-explored modular paradigm of using VLMs as frozen, zero-shot reasoners. Prior work in this area focused on auxiliary tasks like safety-aware decision verification for perception and planning modules \cite{wang2023empowering} or hard-example mining \cite{hard_case}. The most closely related work to ours is VLM-AD \cite{VLM-AD} which leverages VLM-generated reasoning text to construct an auxiliary classification loss.  Another line of work explores aligning BEV features with LM-derived expectation embeddings (ALP) and extending this alignment to ego-vehicle query features for planning (SLP) via contrastive learning \cite{VLP}. While this paradigm enriches perception and planning with semantically meaningful representations, it remains an auxiliary training signal for feature refinement, rather than a direct mechanism for aligning autonomous driving decisions with human preferences. OmniDrive \cite{OmniDrive} generates a counterfactual Q\&A dataset by simulating alternative trajectories and prompting GPT-4 for safety reasoning, offering denser supervision for perception and planning but focusing on dataset quality rather than preference alignment.
Our work explores a novel and complementary direction focusing on simpler and more interpretable forms of supervision: we investigate direct model conditioning on discrete HLAs and, most critically, the generation of explicit preference pairs for human preference alignment. 
This approach is simpler, as it avoids hand-crafting intricate loss functions, and more interpretable, as the quality of the VLM's discrete outputs—a chosen action or a preference—can be directly evaluated.

\textbf{Preference Alignment:} The standard paradigm for aligning models with human preferences is Reinforcement Learning from Human Feedback (RLHF) \cite{ouyang2022training}\cite{christiano2017deep}. While powerful, this multi-stage process of reward modeling and reinforcement learning is notoriously complex and can suffer from training instability. To avoid these issues, our work employs direct preference optimization, a more direct and stable offline method that optimizes a policy via a simple classification loss on preference pairs \cite{DPO}.
DPO has recently been applied to embodied agent tasks, such as motion forecasting with implicit preferences from heuristics \cite{tian2025direct} and robotic manipulation \cite{VTLA}. Our work introduces a fundamentally different approach for generating the supervisory signal. We leverage a VLM in a zero-shot capacity to generate explicit, semantically-grounded preference pairs based on its commonsense understanding of the scene. 

\section{Methodology} 
\label{sec:methodology}

\subsection{Generative Motion Forecasting Model}
\begin{figure}[t] 
    \centering
    \includegraphics[scale=0.21]{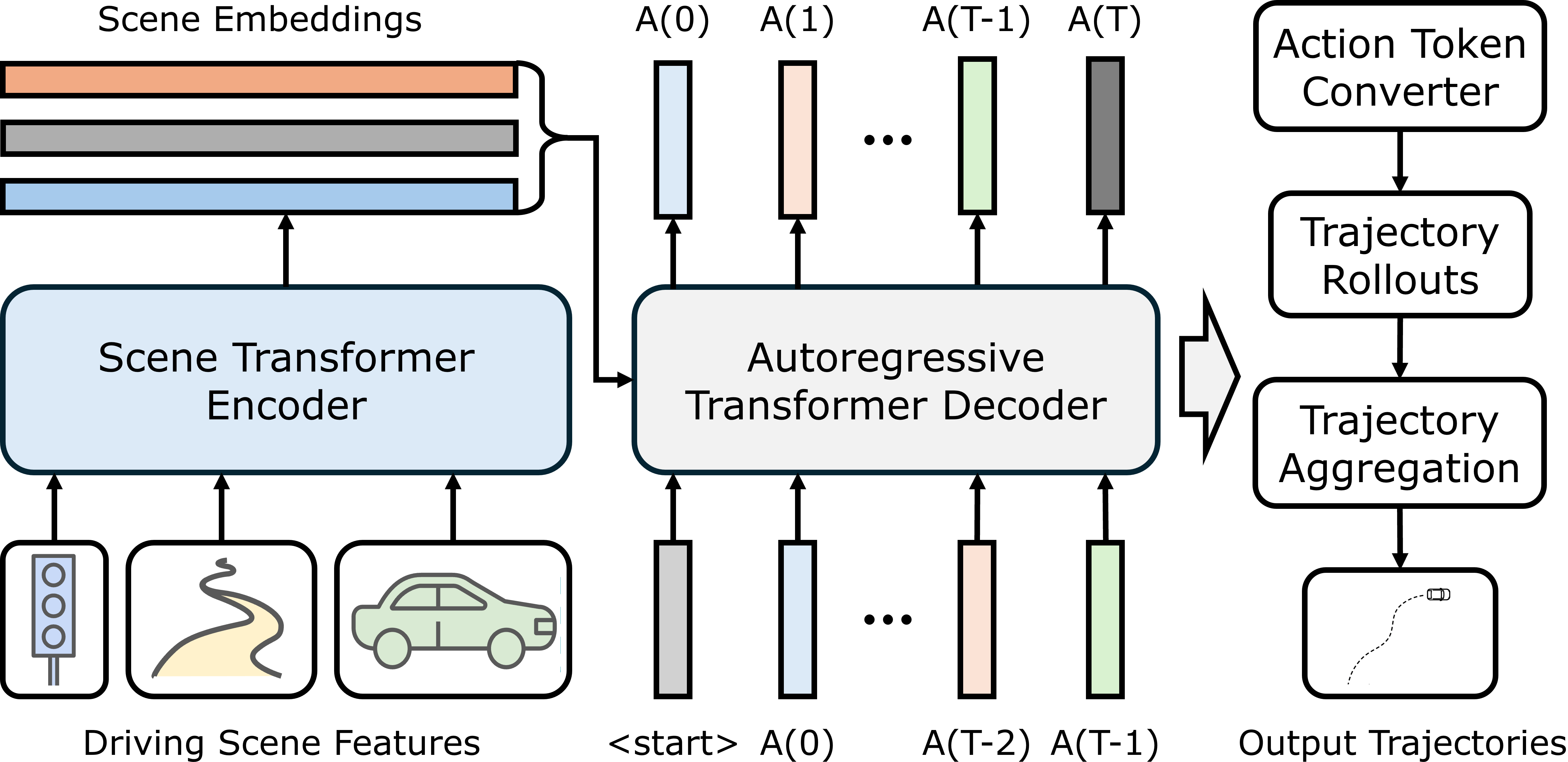}
    \caption{The architecture of MotionLM \cite{motionlm}. It adopts an encoder–decoder transformer that takes scene features as input and autoregressively generates discrete action tokens from a predefined vocabulary, followed by postprocessing to convert these tokens into trajectories across multiple modalities.}
    \label{fig:model_config}
\end{figure}

The motion forecasting model in our framework is pretrained on a joint autoregressive prediction task for 8 agents: the ego vehicle and its 7 nearest neighbors. Following the MotionLM paradigm \cite{motionlm}, we adopt an encoder–decoder transformer architecture and formulate motion forecasting as a sequence generation task conditioned on a rich scene context, $\mathbf{S}$, comprising the road graph, traffic light states, and historical agent trajectories (see Fig. \ref{fig:model_config}). The model generates fine-grained sequences of discrete action tokens for each agent from a finite vocabulary. These tokens represent quantized changes in the agent's position, which can be deterministically converted back into a continuous trajectory. By sampling and aggregating multiple trajectories, the model produces a multi-modal set of 12 predictions for each agent.

The pretraining objective is imitation learning, where parameters are optimized to maximize the log-probability of expert-demonstrated trajectories. 
This allows the model to learn a rich implicit distribution over possible futures $P(a^1_1, a^2_1,..., a^8_1, …, a^1_T, a^2_T,..., a^8_T | \mathbf{S})$ capturing complex multi-agent interactions and yielding a foundational representation suitable for diverse downstream tasks. 
Importantly, this generative formulation enables the model to score candidate trajectories by computing their log-probabilities.

During preference finetuning, we restrict the task to single-agent (ego-vehicle) trajectory prediction since the autonomous vehicle is the only agent whose perception data can be directly processed by the VLM.

\subsection{VLM as a Zero-Shot Driving Reasoner} \label{sec:VLM-CoT}

\subsubsection{Multimodal Scene Representation}
\begin{figure*}[t]
    \centering
    \includegraphics[scale=0.77]{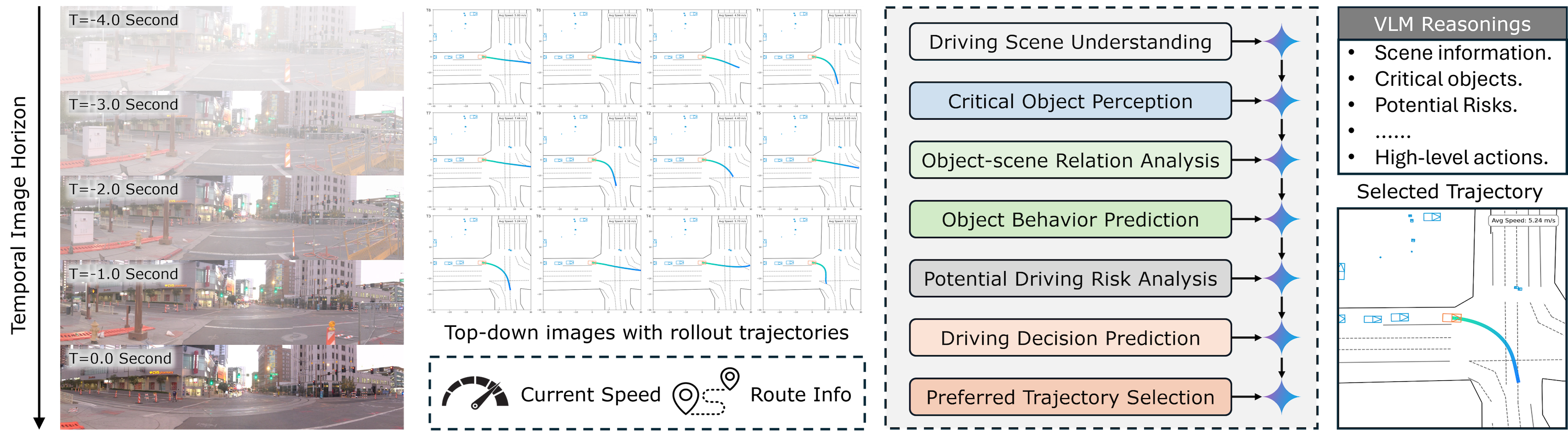}
    \caption{Illustration of the VLM’s Chain-of-Thought (CoT) reasoning process. The VLM takes as input the sequence of image history, top-down view images with sampled trajectories, along with speed and route information. The reasoning flows from perception to driving decision prediction and ultimately to preference-based trajectory selection. The model outputs both the high-level action and the selected trajectory, which serve as finetuning signals.}
    \label{fig:cot_method}
\end{figure*}

To enable the VLM to perform robust context-aware reasoning, we construct a rich multimodal scene representation that provides a holistic view of the driving environment by combining three modalities as illustrated in Fig. \ref{fig:cot_method}.
\begin{itemize}
    \item \textbf{Egocentric and Temporal Vision.} A panoramic image is constructed by stitching the front, front-left, and front-right camera feeds into a wide egocentric view. To capture scene dynamics, we represent this view as a sequence comprising the current frame and four historical frames sampled at 1 Hz.
    \item \textbf{Spatial Reasoning and Visual Grounding.} We generate a top-down, bird's-eye-view (BEV) visualization that renders the local road graph, the positions of other agents, and, most importantly, candidate ego-trajectory predictions from motion forecasting model rollouts. By presenting the candidate trajectories visually within a geometrically precise ego-centric frame, we transform the abstract preference selection task into a more direct visual comparison problem, enabling the VLM to reason about spatial conflicts and path suitability. To present all options in a single input, we generate a BEV image for each of the 12 candidate trajectories and combine them into a composite visualization, then the VLM selects the preferred trajectory by image index. Sec. \ref{finetuning_section} provides details on using this for preference finetuning.
    \item \textbf{Textual Grounding.} To ground the VLM’s reasoning in both long-term goals and immediate dynamics, we add structured text describing key scene features, including the route planner’s navigational command and the ego vehicle’s current speed.
\end{itemize}

\subsubsection{Chain-of-Thought Prompting for Driving Supervision}\label{sec:COT}
To generate high-quality supervisory signals for preference finetuning, we apply Chain-of-Thought prompting, which elicits a structured and interpretable reasoning trace from the VLM. As illustrated in Fig. \ref{fig:cot_method}, this multi-step process guides the model through a hierarchical reasoning cascade—progressing from scene understanding to the final high-level action output and preference trajectory selection.

\begin{enumerate}
    \item \textbf{Scene Understanding:} The designed prompt begins by establishing general context, with the VLM describing the overall driving environment, including road type, prevailing weather, and time of day.
    \item \textbf{Critical Object Perception:} The next step identifies critical objects in the camera image together with their relative positions to the ego vehicle.
    \item \textbf{Object-Scene Relation Identification:} For each critical object, the reasoning process determines interactions with other objects and environmental conditions.
    \item \textbf{Dynamic Object Behavior Prediction:} For each critical moving object detected in previous steps, high-level future actions are predicted from a predefined set.
    \item  \textbf{Potential Risk Analysis:}  The VLM analyzes the scene to identify potential risks or hazards to the ego vehicle and explains why each poses a threat.
    \item \textbf{Driving Decision Prediction:}  Based on the preceding analysis, a high-level action is recommended for the ego vehicle from a predefined set (see Section \ref{sec:HLA}).
    \item \textbf{Preference Trajectory Selection:} Finally, using the full reasoning trace together with the stitched top-down visualization of candidate trajectories, the framework selects the most preferred trajectory for the scenario.
\end{enumerate}

\subsection{VLM-Guided Finetuning Framework} \label{finetuning_section}
To leverage the VLM's supervisory signals, we propose and evaluate two distinct finetuning methodologies: direct supervision on discrete High-Level Actions in Sec. \ref{sec:HLA}, and a more holistic alignment on trajectory-level preferences via direct preference optimization in Sec. \ref{sec:VL-DPO}. 

\subsubsection{Supervision with High-Level Actions (HLAs)} \label{sec:HLA}
To transform the VLM’s free-form reasoning into a structured format, we experiment with three categories of high-level action vocabularies defined as follows: 

\begin{itemize}
    \item \textbf{Maneuver Action:} This action set provides a holistic semantic label for complete driving behaviors, combining both directional and speed intent. It includes: [Move to stop, Stop to move, Left turn, Right turn, Backup, Left lane change, Right lane change, Remain stopped, U-turn, Pullover, Lane following].
    \item \textbf{Direction Action:} This action set specifies the ego vehicle’s directional intent while omitting explicit speed information. It includes: [Stop, Go straight, Turn left, Turn right, Left lane change, Right lane change].
    \item \textbf{Speed Action:} This action set captures the dynamic aspect of motion by describing the intended change in the ego vehicle’s speed profile. It includes: [Maintain speed, Accelerate, Decelerate, Hard brake].
\end{itemize}
\noindent In our framework, the \textbf{maneuver} category serves as the default action definition. We also explore an alternative formulation where \textbf{direction} and \textbf{speed} actions are combined to represent driving intent. These different definitions are primarily evaluated to investigate whether the choice of action representation influences finetuning performance.

With the HLAs from the VLM, we propose two integration strategies: (i) incorporating them in an auxiliary loss, and (ii) conditioning the motion forecasting model on them as input.

\textbf{HLAs as an Auxiliary Loss.} The first approach treats HLA prediction as an auxiliary task within a multi-task learning framework. The underlying hypothesis is that enforcing the model’s latent representations to be predictive of the VLM’s high-level semantic intent encourages a more structured and context-aware representation space, thereby improving the quality of trajectory prediction. To implement this, we attach a lightweight multi-layer perceptron (MLP) head to the decoder’s final hidden state embeddings. This classification head, followed by a softmax layer, is trained to predict the VLM-provided HLA token using cross-entropy loss. The resulting auxiliary loss is combined with the primary imitation learning loss during finetuning.

\textbf{HLAs as Conditional Input.} The second approach leverages the VLM-provided HLA as a conditional input, framing the problem as goal-conditioned trajectory generation. The hypothesis is that explicitly conditioning on the high-level goal reduces the size of the output space, simplifying prediction and improving accuracy. Concretely, the VLM-generated HLA is encoded as a one-hot vector, projected into the model’s embedding dimension, and concatenated with other scene context features before entering the encoder. This design ensures that the entire model is conditioned on the semantic maneuver it is expected to execute.

\subsubsection{Supervision with VLM Trajectory Selections} \label{sec:VL-DPO}
\begin{algorithm}[t] \label{alg:VL-DPO Loss}
\caption{VL-DPO Loss Computation} 
\SetAlgoNoLine%
$\mathcal{\theta}_{ref} \gets \text{pretrained motion forecasting model} $\; \label{alg:ref_model}
$\mathcal{\theta}_{target} \gets \text{target motion forecasting model} $\; \label{alg:target_model}
$\mathcal{S}^{\mathcal{T}}_{\text{rollout}} \gets \textbf{sampleRollouts}(\mathcal{\theta}_{ref}, \text{inputs}, N_{\text{samples}}) $\; \label{alg:rollout}
$\mathcal{T}_{\text{select}} \gets \textbf{getVLMSelectedTraj}(\mathcal{S}^{\mathcal{T}}_{\text{rollout}}, \text{prompts})$\; \label{alg:vlm_selection}
$\mathcal{S}^{\mathcal{T}}_{\text{unselect}} \gets \mathcal{S}^{\mathcal{T}}_{\text{rollout}} \ \setminus \  \mathcal{T}_{\text{select}}$\; \label{alg:unselect_traj_set}
$\pi^{w}_{ref} \gets \textbf{computeProb}(\mathcal{\theta}_{ref}, \mathcal{T}_{\text{select}})$\; \label{alg:select_prob_start}
$\pi^{w} \gets \textbf{computeProb}(\mathcal{\theta}_{target}, \mathcal{T}_{\text{select}})$\;
$\mathcal{L}_{\text{VL-DPO}} \gets 0$\; \label{alg:select_prob_end}
\For{$\normalfont{\mathcal{T}}^{i}_{\text{unselect}}$ \normalfont{\textbf{in}} $\mathcal{S}^{\mathcal{T}}_{\text{unselect}}$}{ 
    $\pi^{l}_{ref} \gets \textbf{computeProb}(\mathcal{\theta}_{ref}, \mathcal{T}^{i}_{\text{unselect}})$\; \label{alg:unselect_prob_start}
    $\pi^{l} \gets \textbf{computeProb}(\mathcal{\theta}_{target}, \mathcal{T}^{i}_{\text{unselect}})$\; \label{alg:unselect_prob_end}
    $\mathcal{L}_{\text{DPO}} \gets \textbf{DPOLoss}(\pi^{w}_{ref}, \pi^{w}, \pi^{l}_{ref}, \pi^{l})$\; \label{alg:DPO Loss}
    $\mathcal{L}_{\text{VL-DPO}} \gets \mathcal{L}_{\text{VL-DPO}} + \mathcal{L}_{\text{DPO}}$\; \label{alg:VL-DPO Loss start}
}
$N_{\text{pairs}} \gets N_{\text{samples}} - 1$\;
$\mathcal{L}_{\text{VL-DPO}} \gets \mathcal{L}_{\text{VL-DPO}} \ / \ N_{\text{pairs}}$\; \label{alg:VL-DPO Loss end}
$\textbf{return} \ \mathcal{L}_{\text{VL-DPO}}$\; 
\end{algorithm}


The standard next-token prediction objective—central to paradigms such as MotionLM \cite{motionlm}—is effective at capturing locally coherent behaviors but does not guarantee long-term consistency. 
This results in a misalignment, as the model may optimize for trajectory prediction accuracy while neglecting aspects humans value, such as yielding to pedestrians, maintaining safe gaps, or avoiding overly aggressive maneuvers.
Direct preference optimization \cite{DPO} addresses this limitation by training on preference pairs, each consisting of a preferred and an unpreferred behavior. The objective increases the likelihood of the preferred example (in our case, the VLM-selected trajectory) while decreasing the likelihood of the unpreferred ones. Building on this idea, we propose a vision–language-guided DPO (VL-DPO) loss that leverages VLM-generated preference pairs to align motion forecasting models with human preferences as detailed in Alg. \ref{alg:VL-DPO Loss}.
 
%
%

To compute the VL-DPO loss, we use a pretrained reference motion forecasting model, $\theta_{ref}$, which remains frozen during finetuning, along with a target model, $\theta_{target}$, initialized from the same pretrained weights (Lines \ref{alg:ref_model}-\ref{alg:target_model}). Specifically, we first sample $N_{\text{samples}}=12$ rollout trajectories from the reference model, render the corresponding top-down BEV representations with surrounding agents and the roadgraph, and then apply the CoT method described in Sec. \ref{sec:COT} to select the most preferred trajectory (Lines Alg. \ref{alg:rollout}–\ref{alg:vlm_selection}). This yields a single selected trajectory, $\mathcal{T}_{\text{select}}$, and a set of 11 unselected trajectories, $\mathcal{S}^{\mathcal{T}}_{\text{unselect}}$ (Line \ref{alg:unselect_traj_set}). From this, we construct 11 preference pairs, each comparing the VLM-selected trajectory against one of the unselected trajectories: 
\begin{equation}
    \{ (\mathcal{T}_{\text{select}}, \mathcal{T}^{i}_{\text{unselect}}) \mid \mathcal{T}_{\text{unselect}} = \mathcal{T}_{\text{rollout}} \setminus \mathcal{T}_{\text{select}} \}.
\end{equation}
For each preference pair, the DPO loss requires computing the probabilities, $\pi$, of both the selected and unselected trajectories under the reference and target models (Lines \ref{alg:select_prob_start}–\ref{alg:select_prob_end} and \ref{alg:unselect_prob_start}–\ref{alg:unselect_prob_end} in Alg. \ref{alg:VL-DPO Loss}). DPO loss for a single pair is: 
\begin{equation}
    \mathcal{L}_{\text{DPO}} =
    -   \log \sigma \left( \beta \log \frac{\pi^{w}}{\pi^{w}_{ref}} - \beta \log \frac{\pi^{l}}{\pi^{l}_{ref}} \right),
\end{equation}
where $\sigma$ denotes the sigmoid function and $\beta$ is a scaling parameter that regulates how much the target model is permitted to deviate from the reference model. The final VL-DPO loss for each example is obtained by averaging the DPO losses over all 11 preference pairs (Lines \ref{alg:VL-DPO Loss start}–\ref{alg:VL-DPO Loss end}).

\section{Result and Discussion}
\label{sec:results}
\subsection{Dataset and Model Configurations}
The motion forecasting model is pretrained on a large-scale internal dataset. Our model adopts the encoder–decoder architecture of MotionLM \cite{motionlm}, pretrained for multi-agent prediction and finetuned for single-agent ego forecasting. 
For finetuning and evaluation, we use the Waymo Open End-to-End Driving Dataset (WOD-E2E)\footnote{https://waymo.com/open/challenges/2025/e2e-driving/}, which contains 2,037 training and 479 validation examples. In each example, our model is given a 4-second observation window and tasked with predicting the ego-vehicle trajectory over the next 5 seconds. To ensure compatibility with the pretrained backbone, we augment WOD-E2E with features extracted using proprietary systems, including road graph information, traffic light states, and historical agent trajectories. For the VLM, we use Gemini 2.5 Pro \cite{comanici2025gemini} in its zero-shot setting. Performance is assessed on the post-aggregation (12) trajectories using the human-annotated preference labels based on:
\begin{itemize}
    \item \textbf{Rater Feedback Score (RFS):} The primary metric for human preference alignment. Each example in WOD-E2E provides three human-annotated trajectories ranked from most to least preferred, with raters assigning scores from 0 (worst) to 10 (best). The model’s predictions are then evaluated against these annotated preferences, where higher scores indicate better alignment. Our motion forecasting model outputs 12 trajectories, so we report three RFS
    metrics: RFS (central-mode trajectory\footnote{centrally located trajectory among all predicted trajectories.}), avgRFS (average over all trajectories), and mlRFS (most-likely trajectory).
    \item \textbf{Average Displacement Error (ADE):} A geometric accuracy metric, computed as the mean L2 distance between the predicted trajectory and the target ego-vehicle trajectory at each timestep over a predefined prediction horizon. Lower values indicate higher performance.
\end{itemize}

\begin{table}[t]
\centering
\renewcommand{\arraystretch}{1.2}
\setlength{\tabcolsep}{6pt} 
\begin{tabular}{l|cc}
\toprule
Method &  RFS $\uparrow$ & ADE (5S) [m] $\downarrow$  \\
\midrule
Most Likely Selection      & 7.2279 & 3.1948  \Tstrut\\
VLM Selection     & 7.2970 & 2.9460  \Tstrut\\
\midrule
VLM Selection Gain    & \textcolor{PineGreen}{+0.96\%} & \textcolor{PineGreen}{-7.79\%}  \Tstrut\\
\bottomrule
\end{tabular}
\caption{Comparison of VLM-selected trajectory with the most likely trajectory from baseline model rollouts.}
\label{tab:vlm_selection_eval}
\vspace{-4pt}
\end{table}

\subsection{VLM Reasoning Evaluation}
To evaluate the VLM-generated supervisory signals, we perform both qualitative and quantitative analyses.

\begin{figure}[t] 
    \centering
    \includegraphics[scale=1.03]{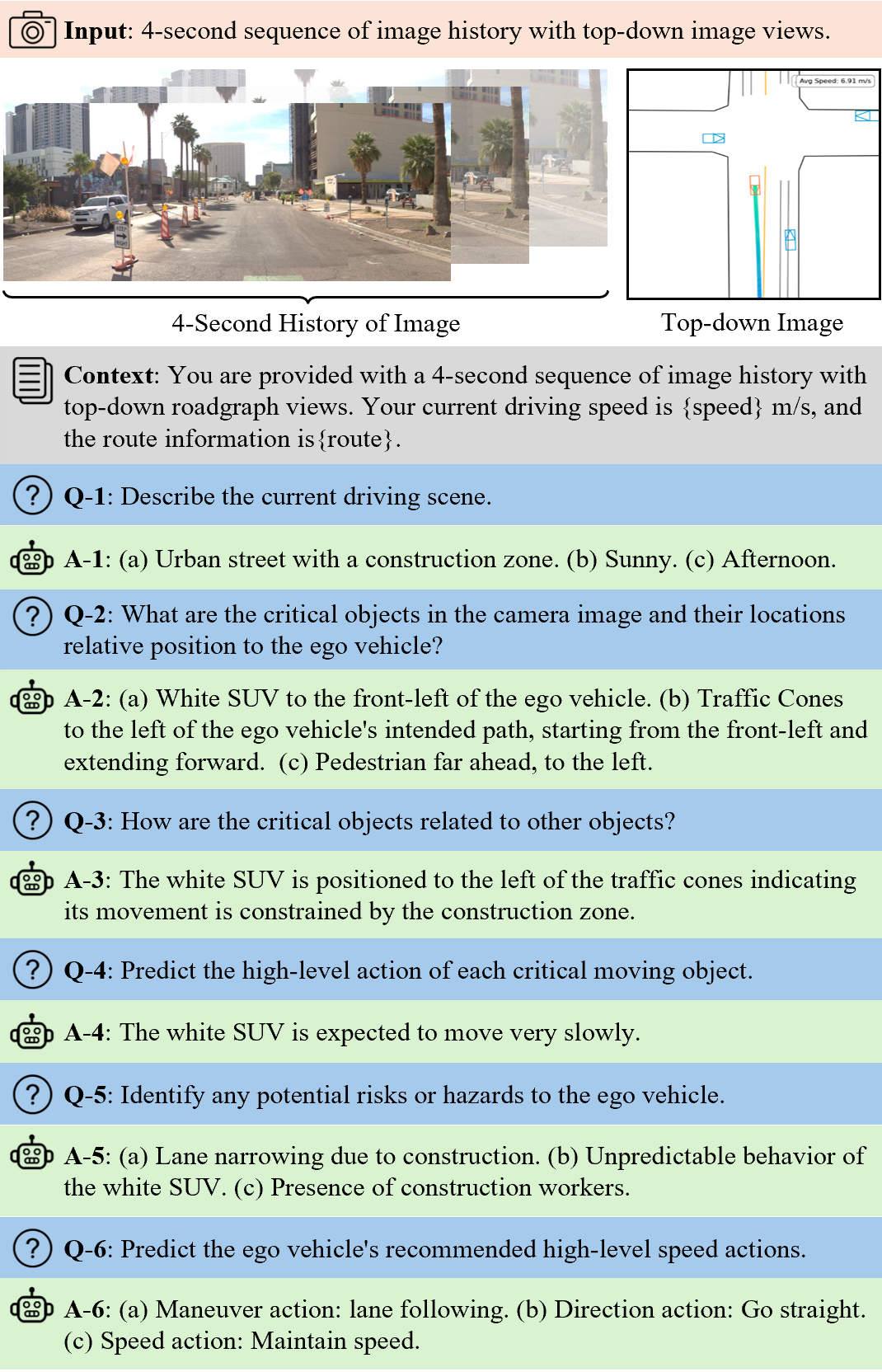}
    \caption{Example of VLM Chain-of-Thought reasoning. For clarity, the full VLM response is simplified, and the trajectory selection step is omitted. }
    \label{fig:vlm_cot_results}
\end{figure}

\textbf{Qualitative Reasoning Analysis.} Fig. \ref{fig:vlm_cot_results} provides a representative example of the VLM's Chain-of-Thought reasoning with simplified prompt questions, intermediate reasoning steps, and final high-level action outputs. 
For brevity, the trajectory selection step (as in Fig. \ref{fig:cot_method}) is omitted.  
In the complex construction-zone scenario of Fig. \ref{fig:vlm_cot_results}, the VLM correctly identifies the street condition and analyzes critical objects in the scene. It successfully reasons about object relations and forecasts the motion of moving agents. Importantly, it highlights potential risks such as lane narrowing due to construction, the possibility of unexpected motion from an oncoming SUV, and the presence of construction workers. Based on this reasoning, the VLM predicts a reasonable maneuver: lane following while going straight and maintaining speed. 

\textbf{Quantitative Selection Performance.} We compare the quality of the VLM's final trajectory choice against the motion prediction model's own probabilistic ranking (i.e., its most likely rollout). As shown in Table \ref{tab:vlm_selection_eval}, the trajectory selected by the VLM is demonstrably better aligned with human preferences, yielding an improved RFS (+0.96\%) and a significant reduction in ADE (–7.79\%). We note that the zero-shot VLM selection in Table~\ref{tab:vlm_selection_eval} yields a much larger improvement in ADE than in RFS. This is a direct consequence of how the two metrics are defined. ADE is computed against the single most-preferred human trajectory, so choosing smoother and safer rollouts translates directly into lower error. RFS, however, is assigned by matching to one of three annotated trajectories with their scores, using trust regions. Once the prediction falls inside the trust region of a given rater trajectory, further geometric improvements (which strongly reduce ADE) do not necessarily change the RFS score, making the gain appear small. Importantly, our finetuning results (Table~\ref{tab:performance_comparison}) demonstrate that this limitation is overcome when VLM selections are turned into preference pairs for DPO training: the pairwise supervision signal targets ranking consistency across all annotated trajectories, leading to much stronger RFS improvements.



\subsection{Finetuning Results}
\begin{table}[t]
\centering
\renewcommand{\arraystretch}{1.05}
\setlength{\tabcolsep}{3.5pt} 
\begin{tabular}{l|cccc}
\toprule
Method & RFS $\uparrow$ & avgRFS $\uparrow$ & mlRFS $\uparrow$ & ADE (5s) $\downarrow$ \\
\midrule
MotionLM \cite{motionlm} & 7.292   & 7.083 & 7.227 &  3.195 \Tstrut\\
Imitation Learning      & 7.844  & 7.861 & 7.855 & 3.075   \Tstrut\\
IL+VL-HLA (Loss)     & 7.841  & 7.799 & 7.779 & 2.825   \Tstrut\\
IL+VL-HLA (Input)    & 8.060  & 8.032 & 8.096 & \textbf{2.715}   \Tstrut\\
\rowcolor{gray!20}
IL+VL-DPO & \textbf{8.163} & \textbf{8.069} & \textbf{8.138} & 2.875 \Tstrut\\
\bottomrule
\end{tabular}
\caption{Performance comparison of MotionLM~\cite{motionlm} baseline and finetuned variants on RFS, avgRFS, mlRFS and 5-s ADE.}
\label{tab:performance_comparison}
\end{table}

\begin{figure}[t] 
    \centering
    \includegraphics[scale=0.94]{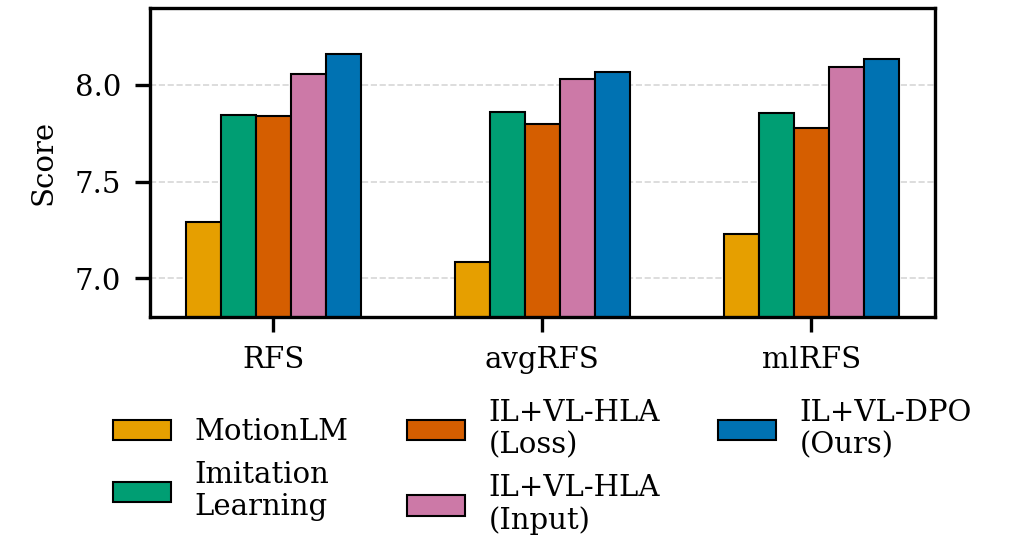}
    \caption{Comparison of  RFS, avgRFS, mlRFS across the finetuning methods. }
    \label{fig:rfs_metrics}
\end{figure}

\begin{figure*}[t] 
    \centering
    \includegraphics[scale=0.8]{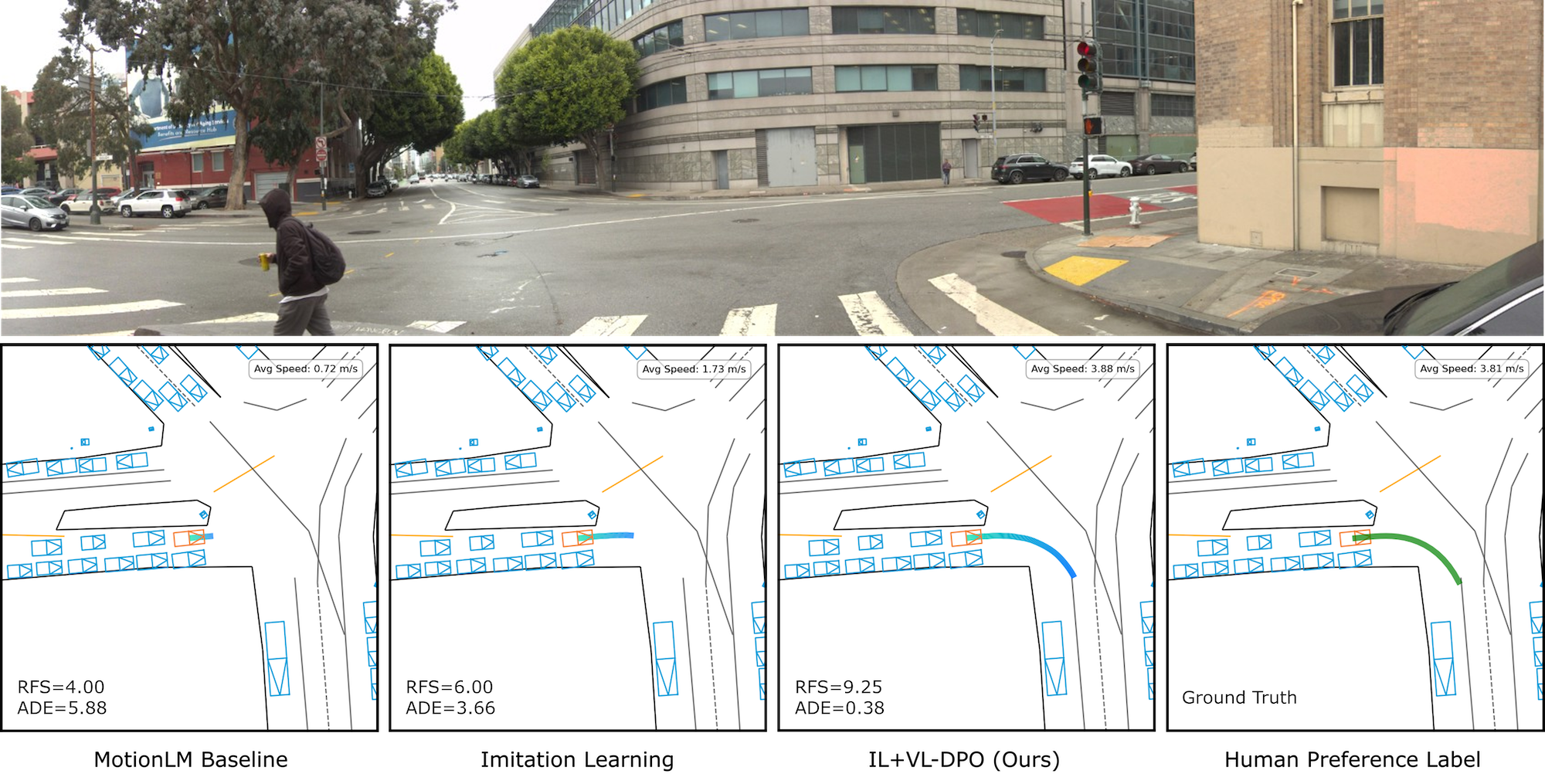}
    \caption{Comparison of central-mode trajectory prediction plots in top-down images from MotionLM \cite{motionlm}, the imitation learning–finetuned model, and our proposed IL+VL-DPO finetuned model, shown alongside the ground-truth human-annotated preference trajectory in an example driving scenario. }
    \label{fig:prediction_example}
\end{figure*}
Having validated our supervisory signals, we now present the main finetuning results, building from a simple baseline to our final proposed method as summarized in Table \ref{tab:performance_comparison}. 

\subsubsection{Imitation Learning (IL)}
The pretrained MotionLM model serves as our initial baseline. As a first step, we finetune it using Imitation Learning (IL) on the single, highest-rated human trajectory in each example. This establishes a very strong baseline, dramatically improving both RFS (avgRFS increases from 7.08 to 7.86) and ADE (from 3.19 to 3.07) and confirming the need for preference alignment.

\subsubsection{Finetuning with High-Level Actions (IL+VL-HLA)}
Next, we explore augmenting the IL baseline with VLM-generated HLAs. We find that using HLAs as a model input (IL+VL-HLA(input)) provides a significant boost, particularly to ADE, achieving our best geometric accuracy of 2.715 in table \ref{tab:performance_comparison}. This shows that providing a coarse semantic anchor answering the question \say{WHAT is the correct maneuver?} helps regularize the model and simplifies the prediction task. In contrast, the auxiliary loss approach (IL+VL-HLA (Loss)) yielded no notable improvement, likely because the gradients from a small classification head are insufficient to meaningfully influence the model's primary representations.

\subsubsection{Vision-Language Guided Preference Tuning (VL-DPO)}

Finally, we apply our primary contribution, augmenting the IL baseline with VLM-generated preference pairs via DPO. This method, IL+VL-DPO, achieves the highest RFS\footnote{As of August 2025, the leading RFS score is 7.986. These scores are not comparable to WOD-E2E leaderboard scores due to differences in pretraining data and finetuning features.}  (Fig. \ref{fig:rfs_metrics}) and competitive ADE. It demonstrates improvements of 11.94\% and 4.07\% in RFS, and reductions of 10.01\% and 6.5\% in ADE as compared to the baseline and imitation learning–finetuned models, respectively. 
This superiority demonstrates that the greatest gains are unlocked by providing a fine-grained, trajectory-level comparative signal that answers the nuanced question: \say{HOW should this maneuver be executed safely and comfortably?}
This approach also holds a critical practical advantage: While the IL+VL-HLA (Input) model achieves a slightly lower ADE, it requires the VLM at inference time, whereas our VL-DPO framework only requires the VLM for offline dataset creation.

To demonstrate the effect of RFS improvements, Fig. \ref{fig:prediction_example} shows an example of central-mode trajectory predictions from the MotionLM baseline, imitation learning–finetuned models, and our IL+VL-DPO. In this scenario, a jaywalking pedestrian has nearly finished crossing the street. While baseline models exhibit conservative behavior, our IL+VL-DPO model generates a more efficient and human-like trajectory, demonstrating its superior situational understanding.



\subsection{Discussion and Ablation Studies}
\subsubsection{Impact of HLA Representation}
The effectiveness of HLA supervision is highly dependent on both the training strategy (auxiliary loss vs. conditional input) and the semantic representation (Maneuver vs. Direction+Speed).
Our ablation (Table \ref{tab:hla_ablation}) reveals a clear conclusion: using HLAs as an auxiliary loss provides an insufficient signal, regardless of the definition.
In contrast, when used as a conditional input, the holistic Maneuver action provides a powerful inductive bias, substantially improving both RFS and ADE.
This finding suggests that for conditioning, a single, comprehensive semantic token is a more effective and direct signal than multiple, disentangled ones.
\begin{table}[t]
\centering
\setlength{\tabcolsep}{3.5pt} 
\renewcommand{\arraystretch}{1.1} 

\begin{tabular}{l|ccc}

\toprule
Method & RFS $\uparrow$ & avgRFS $\uparrow$ & ADE (5s) $\downarrow$ \\
\midrule
Imitation Learning      & 7.8446 & 7.8613 & 3.0752 \\
\midrule

\textcolor{brown}{Maneuver} Loss     & 7.8160 & 7.7994 & 2.8284 \\
\textcolor{NavyBlue}{Direction}+\textcolor{SeaGreen}{Speed} Loss    & 7.8915 & 7.8699 & 2.8655 \\
\textcolor{brown}{Maneuver}+\textcolor{NavyBlue}{Direction}+\textcolor{SeaGreen}{Speed} Loss    & 7.9240 & 7.8178 & 2.8634\\
\midrule

\textcolor{brown}{Maneuver} Input     & \textbf{8.0604} & \textbf{8.0324} & \textbf{2.7232}\\
\textcolor{NavyBlue}{Direction}+\textcolor{SeaGreen}{Speed} Input    & 7.8743 & 7.8962  & 2.9185 \\
\textcolor{brown}{Maneuver}+\textcolor{NavyBlue}{Direction}+\textcolor{SeaGreen}{Speed} Input    & 7.8841 & 7.8733 & 2.8626 \\
\bottomrule
\end{tabular}
\caption{Ablation study on VLM HLAs with different definitions and training strategies, evaluated by RFS and 5-s ADE.}
\label{tab:hla_ablation}
\end{table}

To analyze the sources of performance gains, we conduct ablation studies on our two finetuning strategies.
\subsubsection{Dissecting the Efficacy of DPO}
We now analyze the key components of our best-performing model, IL+VL-DPO, with results in Table \ref{tab:dpo_abalation}.

First, we investigate the relationship between preference optimization and simple imitation learning. 
When evaluated as a standalone method, VL-DPO only improves RFS over the MotionLM baseline, confirming that the VLM's preference signal is effective for alignment even without any human data. However, this comes at the cost of a regression in geometric accuracy (ADE). 
In contrast, we observe that pure Imitation Learning yields higher RFS gains while improving ADE over VL-DPO. This is expected: a direct supervised signal on the top human preference trajectory provides a powerful gradient. 

This leads to our central finding: the combination, IL+VL-DPO, surpasses both standalone methods. This underscores that DPO’s strength is not in replacing imitation learning, but in augmenting it. While IL learns from the single "best" trajectory, DPO leverages all 11 comparisons, teaching the model a nuanced understanding of failure modes that is critical for improving human preference scores.

Next, we compare the data sources for the DPO signal. In Table \ref{tab:dpo_abalation}, IL+VL-DPO model outperforms IL+Preference-DPO (which uses human oracle preferences) in RFS. 
This suggests that the VLM, guided by its structured CoT process, provides a more diverse and informative set of \say{what not to do} signals than the human annotator. 
Human preference data is limited to 3 pairs and can be noisy, whereas our VLM generates up to 11 systematic comparisons per scene, offering a richer and more effective optimization signal.

\begin{table}[t]
\centering
\renewcommand{\arraystretch}{1.2}
\setlength{\tabcolsep}{6pt} 
\begin{tabular}{l|ccc}
\toprule
Method & RFS $\uparrow$ & avgRFS $\uparrow$ & ADE (5S) [m] $\downarrow$ \\
\midrule
MotionLM \cite{motionlm}       & 7.2926 & 7.0839  & 3.1121 \Tstrut\\
VL-DPO only       & 7.4432 & 7.3681 & 3.7895 \Tstrut\\
Imitation Learning       & 7.8446 & 7.8613  & 3.0752 \Tstrut\\
IL+Preference-DPO       & 8.0140 & 7.9784  & 2.6689 \Tstrut\\
\rowcolor{gray!20}
IL+VL-DPO (Ours)       & 8.1637 & 8.0694 & 2.8756 \Tstrut\\
\bottomrule
\end{tabular}
\caption{Ablation study of DPO variants in terms of RFS and ADE.}
\label{tab:dpo_abalation}
\end{table}

\section{Conclusion and Future Work}

This work introduces VL-DPO, a vision-language-guided preference finetuning framework, and establishes a clear hierarchy for leveraging VLM-generated signals. While imitation learning on a human-preferred trajectory provides a strong baseline, we show that augmenting it with VLM-generated High-Level Actions yields further improvements by providing a coarse semantic anchor on \say{WHAT} maneuver to execute. However, our primary finding is that the greatest gains are unlocked by the fine-grained, comparative feedback of VL-DPO, which addresses the critical question of \say{HOW} a maneuver should be performed safely and comfortably.

These findings lead to two powerful conclusions. \textbf{Methodologically}, the most effective approach is a  combination of imitation learning on high-quality positive examples and DPO on a rich set of comparative examples. This dual signal teaches the model both what to do and, critically, the nuanced details of what to avoid. \textbf{Architecturally}, our results validate a modular paradigm where a frozen zero-shot VLM provides state-of-the-art alignment for a specialized motion model. 

\section*{ACKNOWLEDGMENT}
We thank Kate Tolstaya, Kratarth Goel, and Neerja Thakkar for their valuable contributions to this work.
\bibliographystyle{IEEEtran}
\bibliography{bibliography.bib}

@inproceedings{motionlm,
  title={Motionlm: Multi-agent motion forecasting as language modeling},
  author={Seff, Ari and Cera, Brian and Chen, Dian and Ng, Mason and Zhou, Aurick and Nayakanti, Nigamaa and Refaat, Khaled S and Al-Rfou, Rami and Sapp, Benjamin},
  booktitle={Proceedings of the IEEE/CVF International Conference on Computer Vision},
  pages={8579--8590},
  year={2023}
}

@INPROCEEDINGS{wayformer,
  author={Nayakanti, Nigamaa and Al-Rfou, Rami and Zhou, Aurick and Goel, Kratarth and Refaat, Khaled S. and Sapp, Benjamin},
  booktitle={ICRA}, 
  title={Wayformer: Motion Forecasting via Simple \& Efficient Attention Networks}, 
  year={2023},
  volume={},
  number={},
  doi={10.1109/ICRA48891.2023.10160609}}

@inproceedings{MultiPath++,
  author    = {Balakrishnan Varadarajan and
               Ahmed Hefny and
               Avikalp Srivastava and
               Khaled S. Refaat and
               Nigamaa Nayakanti and
               Andre Cornman and
               Kan Chen and
               Bertrand Douillard and
               Chi{-}Pang Lam and
               Dragomir Anguelov and
               Benjamin Sapp},
  title     = {MultiPath++: Efficient Information Fusion and Trajectory Aggregation
               for Behavior Prediction},
  booktitle = {ICRA},
  year      = {2022}
}

@inproceedings{DriveLM,
  title={Drivelm: Driving with graph visual question answering},
  author={Sima, Chonghao and Renz, Katrin and Chitta, Kashyap and Chen, Li and Zhang, Hanxue and Xie, Chengen and Bei{\ss}wenger, Jens and Luo, Ping and Geiger, Andreas and Li, Hongyang},
  booktitle={European conference on computer vision},
  year={2024},
  organization={Springer}
}

@inproceedings{
DriveVLM,
title={Drive{VLM}: The Convergence of Autonomous Driving and Large Vision-Language Models},
author={Xiaoyu Tian and Junru Gu and Bailin Li and Yicheng Liu and Yang Wang and Zhiyong Zhao and Kun Zhan and Peng Jia and XianPeng Lang and Hang Zhao},
booktitle={8th Annual Conference on Robot Learning},
year={2024},
}

@article{WiseAD,
  title={Wisead: Knowledge augmented end-to-end autonomous driving with vision-language model},
  author={Zhang, Songyan and Huang, Wenhui and Gao, Zihui and Chen, Hao and Lv, Chen},
  journal={arXiv preprint arXiv:2412.09951},
  year={2024}
}

@article{
EMMA,
title={{EMMA}: End-to-End Multimodal Model for Autonomous Driving},
author={Jyh-Jing Hwang and Runsheng Xu and Hubert Lin and Wei-Chih Hung and Jingwei Ji and Kristy Choi and Di Huang and Tong He and Paul Covington and Benjamin Sapp and Yin Zhou and James Guo and Dragomir Anguelov and Mingxing Tan},
journal={Transactions on Machine Learning Research},
issn={2835-8856},
year={2025},
note={}
}

@article{OpenDriveVLA,
  title={Opendrivevla: Towards end-to-end autonomous driving with large vision language action model},
  author={Zhou, Xingcheng and Han, Xuyuan and Yang, Feng and Ma, Yunpu and Knoll, Alois C},
  journal={arXiv preprint arXiv:2503.23463},
  year={2025}
}

@inproceedings{RT2,
  title={Rt-2: Vision-language-action models transfer web knowledge to robotic control},
  author={Zitkovich, Brianna and Yu, Tianhe and Xu, Sichun and Xu, Peng and Xiao, Ted and Xia, Fei and Wu, Jialin and Wohlhart, Paul and Welker, Stefan and Wahid, Ayzaan and others},
  booktitle={Conference on Robot Learning},
  year={2023},
}

@article{InstructVLA,
  title={InstructVLA: Vision-Language-Action Instruction Tuning from Understanding to Manipulation},
  author={Yang, Shuai and Li, Hao and Chen, Yilun and Wang, Bin and Tian, Yang and Wang, Tai and Wang, Hanqing and Zhao, Feng and Liao, Yiyi and Pang, Jiangmiao},
  journal={arXiv preprint arXiv:2507.17520},
  year={2025}
}

@article{ChatVLA,
  title={Chatvla: Unified multimodal understanding and robot control with vision-language-action model},
  author={Zhou, Zhongyi and Zhu, Yichen and Zhu, Minjie and Wen, Junjie and Liu, Ning and Xu, Zhiyuan and Meng, Weibin and Cheng, Ran and Peng, Yaxin and Shen, Chaomin and others},
  journal={arXiv preprint arXiv:2502.14420},
  year={2025}
}

@inproceedings{mtp,
  title={Multimodal trajectory predictions for autonomous driving using deep convolutional networks},
  author={Cui, Henggang and Radosavljevic, Vladan and Chou, Fang-Chieh and Lin, Tsung-Han and Nguyen, Thi and Huang, Tzu-Kuo and Schneider, Jeff and Djuric, Nemanja},
  booktitle={IEEE International Conference on Robotics and Automation (ICRA)},
  year={2019},
}

@inproceedings{rules_of_road,
  title={Rules of the road: Predicting driving behavior with a convolutional model of semantic interactions},
  author={Hong, Joey and Sapp, Benjamin and Philbin, James},
  booktitle={CVPR},
  year={2019}
}

@inproceedings{multipath,
  title={MultiPath: Multiple Probabilistic Anchor Trajectory Hypotheses for Behavior Prediction},
  author={Chai, Yuning and Sapp, Benjamin and Bansal, Mayank and Anguelov, Dragomir},
  booktitle={Conference on Robot Learning},
  pages={86--99},
  year={2020},
}

@inproceedings{spagnn,
  title={Spagnn: Spatially-aware graph neural networks for relational behavior forecasting from sensor data},
  author={Casas, Sergio and Gulino, Cole and Liao, Renjie and Urtasun, Raquel},
  booktitle={ICRA},
  year={2020}
}

@inproceedings{lanegcn,
  title={Learning lane graph representations for motion forecasting},
  author={Liang, Ming and Yang, Bin and Hu, Rui and Chen, Yun and Liao, Renjie and Feng, Song and Urtasun, Raquel},
  booktitle={European Conference on Computer Vision},
  year={2020},
}

@article{MTR,
  title={Motion Transformer with Global Intention Localization and Local Movement Refinement},
  author={Shi, Shaoshuai and Jiang, Li and Dai, Dengxin and Schiele, Bernt},
  journal={Advances in Neural Information Processing Systems},
  year={2022}
}

@article{hdgt,
  title={HDGT: Heterogeneous Driving Graph Transformer for Multi-Agent Trajectory Prediction via Scene Encoding},
  author={Jia, Xiaosong and Wu, Penghao and Chen, Li and Li, Hongyang and Liu, Yu and Yan, Junchi},
  journal={CoRL},
  year={2022}
}

@INPROCEEDINGS{e2e_multi-modal,
  author={Prakash, Aditya and Chitta, Kashyap and Geiger, Andreas},
  booktitle={2021 IEEE/CVF Conference on Computer Vision and Pattern Recognition (CVPR)}, 
  title={Multi-Modal Fusion Transformer for End-to-End Autonomous Driving}, 
  year={2021},
  volume={},
  number={},
  doi={10.1109/CVPR46437.2021.00700}}

@INPROCEEDINGS{VAD,
  author={Jiang, Bo and Chen, Shaoyu and Xu, Qing and Liao, Bencheng and Chen, Jiajie and Zhou, Helong and Zhang, Qian and Liu, Wenyu and Huang, Chang and Wang, Xinggang},
  booktitle={ICCV}, 
  title={VAD: Vectorized Scene Representation for Efficient Autonomous Driving}, 
  year={2023},
  volume={},
  number={},
  pages={8306-8316},
  doi={10.1109/ICCV51070.2023.00766}}

@inproceedings{UniAD,
  title={Planning-oriented autonomous driving},
  author={Hu, Yihan and Yang, Jiazhi and Chen, Li and Li, Keyu and Sima, Chonghao and Zhu, Xizhou and Chai, Siqi and Du, Senyao and Lin, Tianwei and Wang, Wenhai and others},
  booktitle={CVPR},
  year={2023}
}

@INPROCEEDINGS{hard_case,
  author={Yang, Yi and Zhang, Qingwen and Ikemura, Kei and Batool, Nazre and Folkesson, John},
  booktitle={2024 IEEE Intelligent Vehicles Symposium (IV)}, 
  title={Hard Cases Detection in Motion Prediction by Vision-Language Foundation Models}, 
  year={2024},
  volume={},
  number={},
  doi={10.1109/IV55156.2024.10588694}}

@inproceedings{DiMa,
  title={Distilling multi-modal large language models for autonomous driving},
  author={Hegde, Deepti and Yasarla, Rajeev and Cai, Hong and Han, Shizhong and Bhattacharyya, Apratim and Mahajan, Shweta and Liu, Litian and Garrepalli, Risheek and Patel, Vishal M and Porikli, Fatih},
  booktitle={Proceedings of the Computer Vision and Pattern Recognition Conference},
  year={2025}
}

@article{VDT-Auto,
  title={Vdt-auto: End-to-end autonomous driving with vlm-guided diffusion transformers},
  author={Guo, Ziang and Gubernatorov, Konstantin and Asfaw, Selamawit and Yagudin, Zakhar and Tsetserukou, Dzmitry},
  journal={arXiv preprint arXiv:2502.20108},
  year={2025}
}

@inproceedings{
wang2023empowering,
title={Empowering Autonomous Driving with Large Language Models: A Safety Perspective},
author={Yixuan Wang and Ruochen Jiao and Simon Sinong Zhan and Chengtian Lang and Chao Huang and Zhaoran Wang and Zhuoran Yang and Qi Zhu},
booktitle={ICLR 2024 Workshop on Large Language Model (LLM) Agents},
year={2024},
}

@article{VLM-AD,
  title={Vlm-ad: End-to-end autonomous driving through vision-language model supervision},
  author={Xu, Yi and Hu, Yuxin and Zhang, Zaiwei and Meyer, Gregory P and Mustikovela, Siva Karthik and Srinivasa, Siddhartha and Wolff, Eric M and Huang, Xin},
  journal={arXiv preprint arXiv:2412.14446},
  year={2024}
}

@article{VTLA,
  title={Vtla: Vision-tactile-language-action model with preference learning for insertion manipulation},
  author={Zhang, Chaofan and Hao, Peng and Cao, Xiaoge and Hao, Xiaoshuai and Cui, Shaowei and Wang, Shuo},
  journal={arXiv preprint arXiv:2505.09577},
  year={2025}
}

@inproceedings{
tian2025direct,
title={Direct Post-Training Preference Alignment for Multi-Agent Motion Generation Model Using Implicit Feedback from Pre-training Demonstrations},
author={Thomas Tian and Kratarth Goel},
booktitle={The Thirteenth International Conference on Learning Representations},
year={2025},
}

@inproceedings{
DPO,
title={Direct Preference Optimization: Your Language Model is Secretly a Reward Model},
author={Rafael Rafailov and Archit Sharma and Eric Mitchell and Christopher D Manning and Stefano Ermon and Chelsea Finn},
booktitle={Thirty-seventh Conference on Neural Information Processing Systems},
year={2023},
}

@article{comanici2025gemini,
  title={Gemini 2.5: Pushing the frontier with advanced reasoning, multimodality, long context, and next generation agentic capabilities},
  author={Comanici, Gheorghe and Bieber, Eric and Schaekermann, Mike and Pasupat, Ice and Sachdeva, Noveen and Dhillon, Inderjit and Blistein, Marcel and Ram, Ori and Zhang, Dan and Rosen, Evan and others},
  journal={arXiv preprint arXiv:2507.06261},
  year={2025}
}

@article{ouyang2022training,
  title={Training language models to follow instructions with human feedback},
  author={Ouyang, Long and Wu, Jeffrey and Jiang, Xu and Almeida, Diogo and Wainwright, Carroll and Mishkin, Pamela and Zhang, Chong and Agarwal, Sandhini and Slama, Katarina and Ray, Alex and others},
  journal={Advances in neural information processing systems},
  volume={35},
  year={2022}
}

@article{christiano2017deep,
  title={Deep reinforcement learning from human preferences},
  author={Christiano, Paul F and Leike, Jan and Brown, Tom and Martic, Miljan and Legg, Shane and Amodei, Dario},
  journal={Advances in neural information processing systems},
  volume={30},
  year={2017}
}

@inproceedings{VLP,
  title={Vlp: Vision language planning for autonomous driving},
  author={Pan, Chenbin and Yaman, Burhaneddin and Nesti, Tommaso and Mallik, Abhirup and Allievi, Alessandro G and Velipasalar, Senem and Ren, Liu},
  booktitle={CVPR},
  year={2024}
}

@inproceedings{OmniDrive,
  title={Omnidrive: A holistic vision-language dataset for autonomous driving with counterfactual reasoning},
  author={Wang, Shihao and Yu, Zhiding and Jiang, Xiaohui and Lan, Shiyi and Shi, Min and Chang, Nadine and Kautz, Jan and Li, Ying and Alvarez, Jose M},
  booktitle={Proceedings of the Computer Vision and Pattern Recognition Conference},
  year={2025}
}

@article{baniodeh2025scaling,
  title={Scaling Laws of Motion Forecasting and Planning--A Technical Report},
  author={Baniodeh, Mustafa and Goel, Kratarth and Ettinger, Scott and Fuertes, Carlos and Seff, Ari and Shen, Tim and Gulino, Cole and Yang, Chenjie and Jerfel, Ghassen and Choe, Dokook and others},
  journal={arXiv preprint arXiv:2506.08228},
  year={2025}
}

@inproceedings{
huang2024drivegpt,
title={Drive{GPT}: Scaling Autoregressive Behavior Models for Driving},
author={Xin Huang and Eric M Wolff and Paul Vernaza and Tung Phan-Minh and Hongge Chen and David S Hayden and Mark Edmonds and Brian Pierce and Xinxin Chen and Pratik Elias Jacob and Xiaobai Chen and Chingiz Tairbekov and Pratik Agarwal and Tianshi Gao and Yuning Chai and Siddhartha Srinivasa},
booktitle={Forty-second International Conference on Machine Learning},
year={2025},
}
\end{document}